\documentclass[letterpaper,10 pt,conference]{ieeeconf}
\usepackage{graphics}
\usepackage{graphicx}
\usepackage{tabularx}
\usepackage{multirow}
\usepackage{cite}
\usepackage{subfigure}
\usepackage{amsmath}
\usepackage{amssymb}
\usepackage{bm}
\usepackage{cite}
\usepackage{scalerel}

\IEEEoverridecommandlockouts
\begin{document}

\title{\LARGE \bf Transfer Learning in Brain-Computer Interfaces with Adversarial Variational Autoencoders}

\author{Ozan \"{O}zdenizci$^{\star}$, Ye Wang$^{\dagger}$, Toshiaki Koike-Akino$^{\dagger}$, Deniz Erdo\u{g}mu\c{s}$^{\star}$%
\thanks{$^{\star}$Department of Electrical and Computer Engineering, Northeastern University, Boston, MA, USA, E-mail: \{oozdenizci, erdogmus\}@ece.neu.edu.}%
\thanks{$^{\dagger}$Mitsubishi Electric Research Laboratories (MERL), Cambridge, MA, USA. E-mail: \{yewang, koike\}@merl.com. }%
\thanks{D. Erdo\u{g}mu\c{s} is partially supported by NSF (IIS-1149570, CNS-1544895), NIDLRR (90RE5017-02-01), and NIH (R01DC009834). O. \"{O}zdenizci was an intern at MERL during this work.}%
}
\maketitle


\begin{abstract}
We introduce adversarial neural networks for representation learning as a novel approach to transfer learning in brain-computer interfaces (BCIs). The proposed approach aims to learn subject-invariant representations by simultaneously training a conditional variational autoencoder (cVAE) and an adversarial network. We use shallow convolutional architectures to realize the cVAE, and the learned encoder is transferred to extract subject-invariant features from unseen BCI users' data for decoding. We demonstrate a proof-of-concept of our approach based on analyses of electroencephalographic (EEG) data recorded during a motor imagery BCI experiment.
\end{abstract}

\begin{keywords}
representation learning, transfer learning, adversarial networks, variational autoencoders, convolutional neural networks, EEG, brain-computer interfaces.
\end{keywords}

\section{Introduction}

Transfer learning often describes an approach to discover and exploit some shared structure in the data that is invariant across data sets. In the context of brain-computer interfaces (BCIs), where the aim is to provide a direct neural communication and control channel for individuals, e.g., with severe neuromuscular disorders, the concept of transfer learning gains significant interest given its potential benefit in reducing BCI system calibration times by exploiting neural data recorded from other subjects. Given the limited data collection times under adequate concentration and consciousness with patients, this becomes essential for a potential patient end-user of the BCI system. Several pieces of work in this domain aim to find neural features (representations) that are invariant across subjects or sessions to calibrate BCIs~\cite{Krauledat:2008,Kang:2009,Samek:2013}, or learn a structure for the set of decision rules and how they differ across subjects and sessions~\cite{Alamgir:2010,Jayaram:2016}.

Going beyond neural interfaces, significant progress was recently achieved in domain transfer learning by adversarially censored invariant representations within the growing field of deep learning in computer vision and image processing~\cite{Edwards:2015, Makhzani:2015, Mathieu:2016, lample2017fader, Creswell:2017, Tzeng:2017, Shen:2017, Wang:2018}. These methods rely on learning generative models of the data that allow synthesis of data samples from latent representations, which can be achieved with variational autoencoders (VAEs)~\cite{Kingma:2013} for unsupervised feature learning, or generative adversarial networks (GANs)~\cite{Goodfellow:2014}, where the supervision is alleviated by penalizing inaccurate samples using an adversarial game. Consistently, these are trained with adversarial censoring to learn representations that are aimed to be independent from some nuisance variables (e.g., a representative variable for factors of variations across data sets). In the light of these recent work, we introduce this progress in adversarial representation learning as a novel approach for transfer learning in BCIs.

Various aspects of deep convolutional neural networks (CNNs) in computer vision have been already introduced to extract features for task-specific decoding in electoencephalogram (EEG) based BCIs~\cite{Lawhern:2016,Schirrmeister:2017}, as well as for recent attempts to learn deep generative models for EEG~\cite{Bashivan:2015,Luo:2018,Hartmann:2018}. In the present study, we extend these lines of work and propose a transfer learning approach for BCIs based on the exploitation of adversarial training for subject-invariant representation learning. Particularly, the proposed approach~\cite{lample2017fader, Wang:2018} aims to learn subject-invariant representations by simultaneously training a conditional VAE and an adversarial network to enforce invariance of the learned data representations with respect to subject identity.
This adversarial training procedure, with VAEs based on CNN architectures, yields data representations that work as features that are disentangled from subject-specific nuisance variations, which enables decoding for unseen BCI subjects.
Our results demonstrate the advantage of this approach with a proof-of-concept based on analyses of EEG data recorded from 103 subjects during a motor imagery BCI experiment.

\section{Methods}

\subsection{Notation}

Let $\{(\bm{X}_i^s,y_i^s)\}_{i=1}^{n_s}$ denote the data set for subject $s$ consisting of $n_s$ trials, where $\bm{X}_i^s\in\mathbb{R}^{C \times T}$ is the raw EEG data at trial $i$ recorded from $C$ channels for $T$ discretized time samples, and $y_i^s$ is the corresponding class label from a set of $L$ class labels. In a subject-to-subject transfer learning problem, the aim is to learn a parametric encoder $q_{\phi}(\bm{z}\vert\bm{X})$ which can be generalized across subjects, and extracts latent representations $\bm{z}\in\mathbb{R}^{d_{\bm{z}}}$ from the data $\bm{X}$ that are useful in discriminating different tasks or brain states indicated by their corresponding class labels $y$. Accordingly, let $\bm{s}$ denote the one-hot encoded subject identifier vector for subject $s$ (i.e., an $S$-dimensional vector with a value of 1 at the $s$'th index and zero in other indices), which represents the nuisance variable in our adversarial representation learning frameworks that $\bm{z}$ will be enforced to be independent of.

\begin{figure*}[h!]
      \centering
	\includegraphics[clip, trim=0 8.3cm 0 0, width=\textwidth]{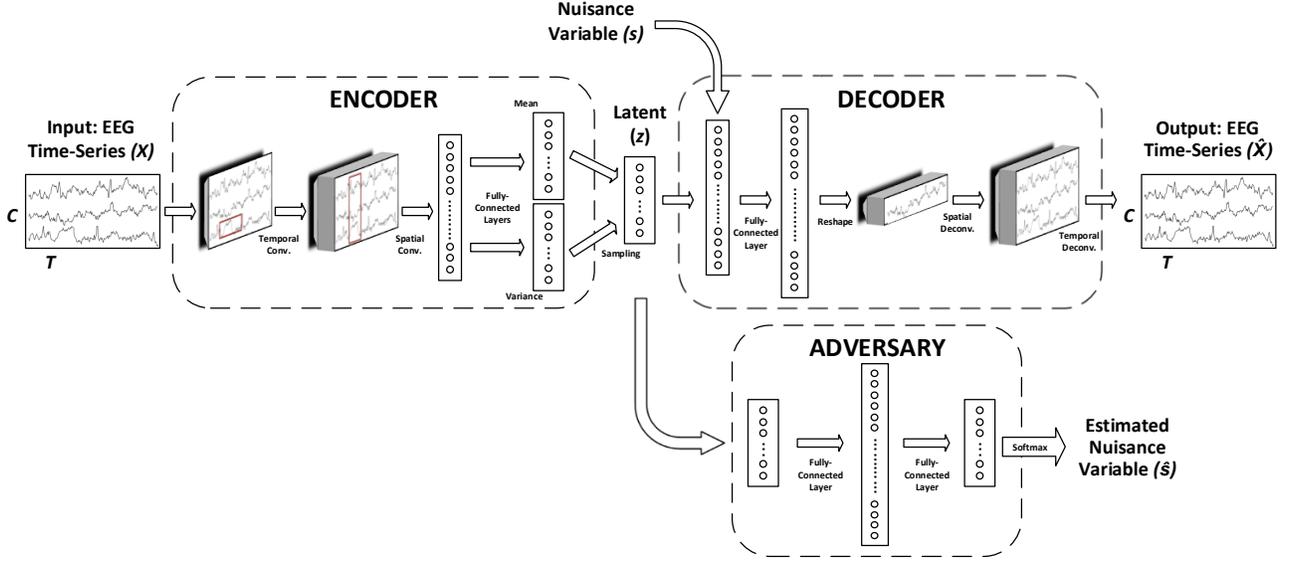}
	\vspace{-1.1cm}
	\caption{Adversarial cVAE (A-cVAE) architecture with a stochastic encoder and a deterministic decoder with conditioning on $\bm{s}$. A-cVAE is trained to minimize the loss function in Eq.~(\ref{eq:acvae}), while the adversary is also individualy trained to minimize its softmax cross-entropy loss. Parameter updates were performed alternatingly among the A-cVAE and the adversary once per batch.}
	\label{fig:acvae}
	\vspace{-0.1cm}
\end{figure*}

\subsection{Conditional Variational Autoencoder (cVAE)}
\label{sec:cvae}

VAEs~\cite{Kingma:2013} learn a generative model as a pair of encoder and decoder networks. The encoder learns a latent representation $\bm{z}$ from the data $\bm{X}$, while the decoder aims to reconstruct the data $\bm{X}$ from the learned representation $\bm{z}$. In this variational framework the encoder is stochastic, meaning that the decoder uses a learned posterior $q_{\phi}(\bm{z}\vert\bm{X}) \sim \mathcal{N}(\bm{\mu_z},\bm{\sigma_z})$, whose parameters are given by the encoder network. The decoder is provided with samples from this posterior distribution as input $\bm{z}$.

In the conditional VAE (cVAE) framework~\cite{Sohn:2015}, the decoder is conditioned on a nuisance variable $\bm{s}$ as an additional input besides $\bm{z}$, and the encoder is expected to learn representations $\bm{z}$ that are invariant of $\bm{s}$, since $\bm{s}$ is already given as input to the decoder. The loss function to be minimized in this cVAE framework, which is also known as the evidence lower bound (ELBO), is given by:
\begin{equation}
\begin{split}
\mathcal{L}_{\scaleto{\text{cVAE}}{4pt}}(\bm{X}_i^s;\theta,\phi) = & - \mathbb{E}\bigl[ \log p_{\theta}(\bm{X}_i^s \vert \bm{z},\bm{s}_i) \bigr] \\
& + D_\mathrm{KL}\bigl( q_{\phi}(\bm{z}\vert\bm{X}_i^s)\vert\vert p(\bm{z}) \bigr),
\end{split}
\label{eq:cvae}
\end{equation}
where the first term is the reconstruction loss of the decoder, and the second term is the encoder variational posterior loss. This framework implicitly enforces invariance for $\bm{z}$ with respect to $\bm{s}$. However this is known to be not perfectly achieved in practice, which paves the way for adversarial training methods in representation learning \cite{Wang:2018}.

\subsection{Adversarial Conditional VAE (A-cVAE)}
\label{sec:acvae}

In the proposed adversarial cVAE (A-cVAE) framework~\cite{lample2017fader, Wang:2018}, a conditional VAE and an adversary to enforce invariance with respect to $\bm{s}$ (i.e., subject identifiers) are simultaneously trained. Specifically, alongside a cVAE that takes EEG time-series data $\bm{X}$ as input to the encoder and estimates $\hat{\bm{X}}$ at the decoder, an adversary is trained that takes learned representations $\bm{z}$ as input, and estimates $\hat{\bm{s}}$.

We extend Eq.~(\ref{eq:cvae}) to obtain the A-cVAE loss function. For the deterministic decoder, reconstruction loss is determined by the mean squared error of the estimated time-series EEG data. Furthermore, softmax cross-entropy loss of the adversary network is inversely added to the loss function for A-cVAE which is then denoted as:
\begin{equation}
\begin{split}
\mathcal{L}_{\scaleto{\text{A-cVAE}}{4pt}}(\bm{X}_i^s;\theta,\phi,\varPsi) = & \| \bm{X}_i^s - \hat{\bm{X}}_i^s \|^2 \\
& + D_\mathrm{KL}\bigl( q_{\phi}(\bm{z}\vert\bm{X}_i^s)\vert\vert p(\bm{z}) \bigr) \\
& + \lambda \, \mathbb{E}\bigl[ \log q_{\varPsi}(\bm{s}_i \vert \bm{z}) \bigr],
\end{split}
\label{eq:acvae}
\end{equation}
where $\lambda > 0$ is a weight parameter to adjust the impact of adversarial censoring on learned representations. Alternatingly once per batch with A-cVAE parameter updates, the adversary is also individually trained to minimize its softmax cross-entropy loss $\mathcal{L}_{\scaleto{\text{A}}{4pt}}(\bm{z};\varPsi) = \mathbb{E}[-\log q_{\varPsi}(\bm{s}_i \vert \bm{z})]$.

\subsection{Model Architecture and Classifier Training}
\label{sec:archi}

In our implementations, the encoder and decoder have convolutional architectures embedding temporal and spatial filterings motivated by the results achieved with EEGNet~\cite{Lawhern:2016}, Deep ConvNet and Shallow ConvNet~\cite{Schirrmeister:2017}. Parameterization and details of the convolutional cVAE architecture are broadly illustrated in Fig.~\ref{fig:acvae}, and provided in detail in Table~\ref{tab:vae}. The two fully connected layers at the output of the encoder generate two $d_{\bm{z}}$-dimensional parameter vectors $\bm{\mu_z}$ and $\bm{\sigma_z}$, which are then used to sample $\bm{z}$. The nuisance variable vector $\bm{s}$ is then concatenated to the sampled $\bm{z}$ as the input for the decoder. We used temporal convolution kernels of size $W=100$, and spatial convolution kernels of size $C=64$, and a latent vector dimensionality of $d_{\bm{z}}=100$. Adjacent to the cVAE, the adversary is realized as a single hidden layer multilayer perceptron (MLP) with ReLU nonlinearity after the first layer, and we fixed adversarial censoring weight parameter $\lambda=1.0$ in all experiments.

Following adversarial representation learning using a set of training data samples, the encoder is kept static and then using the same training data samples, a classifier is trained that is connected to the output of the encoder. Specifically, all training data samples were again used as input to the static encoder that was previously optimized, and using the obtained parameters at the output of the encoder, a latent vector $\bm{z}$ is sampled which was then used as an input to a classifier. The classifier was also realized as a single hidden layer MLP with ReLU nonlinearity after the first layer. Classifier training was performed to minimize its softmax cross-entropy loss $\mathcal{L}_{\scaleto{\text{C}}{4pt}}(\bm{z};\varOmega) = \mathbb{E}[-\log q_{\varOmega}(y \vert \bm{z})]$. The adversary network had output dimensionality of $S$, and the classifier had an output dimensionality of $L$. Both the adversary and the classifier hidden layers had $100$ nodes.

\renewcommand{\arraystretch}{1.5}
\begin{table}
	\caption{A-cVAE Encoder and Decoder Architectures}
	\label{tab:vae}
	\vspace{-0.3cm}
	\begin{center}
	\begin{tabular}{c|c|c}
		\hline
		Layer & Input Dim. & Operation \\
		\hline
		\multirow{2}{*}{Encoder 1} & $C\times T$ & 40 $\times$ Temporal Conv1D ($1 \times W$) \\
		 & $40\times C \times T$ & BatchNorm + ReLU + Dropout (0.25) \\
		\hline
		\multirow{2}{*}{Encoder 2} & $40\times C \times T$ & 40 $\times$ Spatial Conv1D ($C \times 1$) \\
		 & $40\times 1 \times T$ & BatchNorm + ReLU + Dropout (0.25) \\
		\hline
		\multirow{2}{*}{Encoder 3} & $40\times 1 \times T$ & Reshape (Flatten) \\
		 & $40T \times 1$ &  2 $\times$ Fully-Connected Layers \\
		\hline
		Latent ($\bm{z}$) & $d_{\bm{z}}$ & Sample $\bm{z}$ with estimated parameters \\
		\hline
		\multirow{2}{*}{Decoder 1} & $(d_{\bm{z}}+S)\times 1$ & Fully-Connected Layer \\
		 & $40T \times 1$ & ReLU + Reshape \\
		\hline
		\multirow{2}{*}{Decoder 2} & $40\times 1 \times T$ & 40 $\times$ Spatial Deconv1D ($C \times 1$) \\
		 & $40\times C \times T$ & BatchNorm + ReLU + Dropout (0.25) \\
		\hline
		\multirow{2}{*}{Decoder 3} & $40\times C \times T$ & 40 $\times$ Temporal Deconv1D ($1 \times W$) \\
		 & $C\times T$ & BatchNorm + ReLU + Dropout (0.25) \\
		\hline
	\end{tabular}
	\end{center}
	\vspace{-0.5cm}
\end{table}

\subsection{Dataset and Implementation}

We used the publicly available PhysioNet EEG Motor Movement/Imagery Dataset~\cite{Goldberger:2000}, which was collected using the BCI2000 instrumentation system~\cite{Schalk:2004}. The dataset consists of over 1500 one- and two-minute EEG recordings, obtained from 109 subjects. Throughout the experiments, subjects were placed in front of a computer screen and were instructed to perform cue-based motor execution/imagery tasks while 64-channel EEG were recorded at a sampling rate of 160 Hz. These tasks included executing the movement of the right or left hand, opening and closing of both fists or legs; or just the imagination of these movements. Each trial lasted four-seconds with inter-trial resting periods of same length. At the beginning of the experiments, eyes-open and eyes-closed resting-state EEG were also recorded. Each subject participated in the experiment for a single session.

From this data set, six subjects' data were discarded due to irregular timestamp alignments, resulting in a total of 103 subjects. We used trials that correspond to right and left hand motor imagination to evaluate our proposed approach on a conventional BCI paradigm~\cite{Pfurtscheller:2001}. This resulted in a total of 45 four-second trials per subject, with binary class labels $y_i^s$ corresponding to right or left hand imagery. We randomly selected 13 subjects to hold-out for further across-subjects transfer learning experiments. Using the remaining 90 subjects' data, the networks were trained over a training set of 3240 trials, while validations were performed with the remaining 810 trials including data from all subjects. We implemented all analyses with the Chainer deep learning framework~\cite{Tokui:2015}. Networks were trained with 100 trials per batch for 750 epochs ($\sim$25,000 iterations), and parameter updates were performed once per batch with Adam~\cite{Kingma:2015}.

\section{Experimental Results}

\subsection{EEG Pre-Processing and Model Evaluation}

All subjects' data were epoched into the time-interval where the neural changes induced by motor imagery are emphasized~\cite{Pfurtscheller:2001}. Specifically, from the four second duration, the 1-to-3 seconds interval after the imagery cue onset were extracted to be used in experiments, resulting in a time-series length of $T=320$. Raw EEG data were normalized to have zero mean. Note that this pre-processing statistics (i.e., data mean) is only computed on the training data, and then applied to validation and transfer subjects' data.

We evaluate adversarial representation learning with the following frameworks: (1) A-cVAE, (2) cVAE, (3) adversarially censored VAE without conditioning (A-VAE), (4) basic convolutional encoder (CNN). Implementation of (1) corresponds to the Sections~\ref{sec:acvae} and~\ref{sec:archi}. The approach in (2) is expected to reveal the practical deficiencies of only using decoder conditioning for representation invariance. In that case, we still train an adversary in parallel but do not feed the adversarial loss to the overall training objective, i.e., using Eq.~(\ref{eq:cvae}). Method (3) is expected to reveal the tradeoff between enforcing invariance with an adversary but still preserving enough information in $\bm{z}$ to allow sufficient decoder learning (c.f.~a similar approach in~\cite{Edwards:2015}). This corresponds to using the same objective as A-cVAE, but not providing $\bm{s}$ at the decoder input. Finally, (4) depicts a baseline case that uses the same CNN encoder architecture in combination with an MLP classifier but only trained end-to-end from scratch (via softmax cross-entropy loss for classification) rather first training the encoder within a VAE.

\subsection{Across-Subjects Transfer Learning}

To observe representation invariance, accuracies of the adversary network over 90 subjects after training are presented in Table~\ref{tab:advacc}. In this context, a higher accuracy indicates more subject-specific information remaining in the learned representations $\bm{z}$, which results in better decoding of $\bm{s}$ by the adversary. Therefore a lower adversary accuracy is representative of better invariant representation learning, as observed through the least leakage with A-cVAE.

\begin{figure}
      \centering
	\includegraphics[scale=0.55]{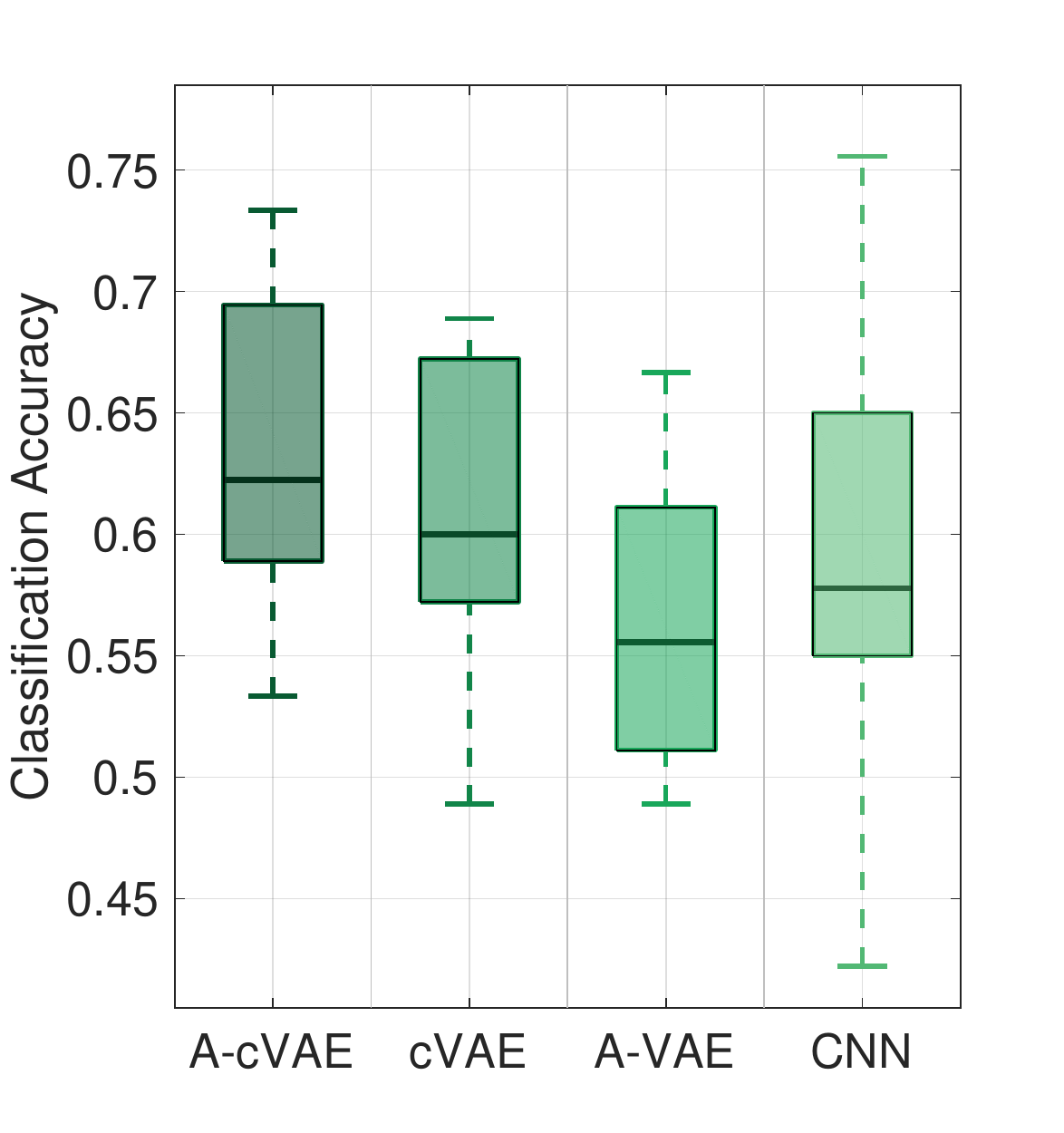}
	\vspace{-0.5cm}
	\caption{Transfer learning classification accuracies for the 13 held-out subjects with learned features by: (1) A-cVAE, (2) cVAE, (3) A-VAE (i.e., no conditioning on $\bm{s}$), (4) CNN as a baseline. Central line mark represents the median across 13 subjects. Upper and lower bounds of the box represents the first and third quartiles. Dashed lines represent the extreme samples.
	Mean accuracies are: (1) 63.8\%, (2) 61.2\%, (3) 56.9\%, (4) 59.8\%.}
	\label{fig:tlacc}
	\vspace{-0.3cm}
\end{figure}

Distributions of transfer learning classification accuracies for the 13 held-out subjects are shown in Fig.~\ref{fig:tlacc}. Using zero subject-specific training or fine-tuning data, we observe accuracies up to $73\%$ with A-cVAE. Consistently with the results in Table~\ref{tab:advacc}, we observe a decrease of accuracies in cVAE and A-VAE with respect to A-cVAE. For baseline CNN, the model tends to memorize the training data without any subject-invariance attempt, resulting in high variation of accuracies across the 13 subjects as intuitively expected.

\renewcommand{\arraystretch}{1.5}
\begin{table}[t]
	\caption{Adversary Accuracies After Model Training}
	\label{tab:advacc}
	\vspace{-0.2cm}
	\begin{center}
	\begin{tabular}{c|c|c|c|c|c}
		\hline
		\multicolumn{3}{c|}{Training Data} & \multicolumn{3}{c}{Validation Data} \\
		\hline
		A-cVAE & cVAE & A-VAE & A-cVAE & cVAE & A-VAE \\
		\hline
		$0.48$ & $0.56$ & $0.68$ & $0.13$ & $0.15$ & $0.21$ \\
		\hline
	\end{tabular}
	\end{center}
	\vspace{-0.5cm}
\end{table}

\section{Discussion}

In this work we introduced adversarial invariant representation learning as a novel approach to transfer learning in BCIs. We revealed that learning subject-invariant representations by adversarial censoring can be a significantly useful tool for subject-transfer learning. We demonstrated an empirical proof-of-concept with EEG data recorded from 103 subjects during a motor imagery BCI experiment.

Hereby, we mainly focused on the results regarding the invariance of representations and the across-subjects transfer learning capability of the models. However the proposed approach can be further extended in the context of semi-supervised transfer learning in BCIs, such as using a short calibration time for fine-tuning and semi-supervised transfer, learning session-invariant representations to reduce user-oriented BCI system calibration times, or learning disentangled representations that exploit adversarial censoring to learn partly subject-invariant, and partly subject-variant representations. We highlight that these frameworks should be of significant interest in the field of neural interfaces.



\begin{thebibliography}{10}
\providecommand{\url}[1]{#1}
\csname url@samestyle\endcsname
\providecommand{\newblock}{\relax}
\providecommand{\bibinfo}[2]{#2}
\providecommand{\BIBentrySTDinterwordspacing}{\spaceskip=0pt\relax}
\providecommand{\BIBentryALTinterwordstretchfactor}{4}
\providecommand{\BIBentryALTinterwordspacing}{\spaceskip=\fontdimen2\font plus
\BIBentryALTinterwordstretchfactor\fontdimen3\font minus
  \fontdimen4\font\relax}
\providecommand{\BIBforeignlanguage}[2]{{%
\expandafter\ifx\csname l@#1\endcsname\relax
\typeout{** WARNING: IEEEtran.bst: No hyphenation pattern has been}%
\typeout{** loaded for the language `#1'. Using the pattern for}%
\typeout{** the default language instead.}%
\else
\language=\csname l@#1\endcsname
\fi
#2}}
\providecommand{\BIBdecl}{\relax}
\BIBdecl

\bibitem{Krauledat:2008}
M.~Krauledat, M.~Tangermann, B.~Blankertz, and K.-R. M{\"u}ller, ``Towards zero
  training for brain-computer interfacing,'' \emph{PloS one}, vol.~3, no.~8, p.
  e2967, 2008.

\bibitem{Kang:2009}
H.~Kang, Y.~Nam, and S.~Choi, ``Composite common spatial pattern for
  subject-to-subject transfer,'' \emph{IEEE Signal Processing Letters},
  vol.~16, no.~8, pp. 683--686, 2009.

\bibitem{Samek:2013}
W.~Samek, F.~C. Meinecke, and K.-R. M{\"u}ller, ``Transferring subspaces
  between subjects in brain--computer interfacing,'' \emph{IEEE Transactions on
  Biomedical Engineering}, vol.~60, no.~8, pp. 2289--2298, 2013.

\bibitem{Alamgir:2010}
M.~Alamgir, M.~Grosse-Wentrup, and Y.~Altun, ``Multitask learning for
  brain-computer interfaces,'' in \emph{Proceedings of the 13th International
  Conference on Artificial Intelligence and Statistics}, 2010, pp. 17--24.

\bibitem{Jayaram:2016}
V.~Jayaram, M.~Alamgir, Y.~Altun, B.~Sch\"{o}lkopf, and M.~Grosse-Wentrup,
  ``Transfer learning in brain-computer interfaces,'' \emph{IEEE Computational
  Intelligence Magazine}, vol.~11, no.~1, pp. 20--31, 2016.

\bibitem{Edwards:2015}
H.~Edwards and A.~Storkey, ``Censoring representations with an adversary,''
  \emph{arXiv preprint arXiv:1511.05897}, 2015.

\bibitem{Makhzani:2015}
A.~Makhzani, J.~Shlens, N.~Jaitly, I.~Goodfellow, and B.~Frey, ``Adversarial
  autoencoders,'' \emph{arXiv preprint arXiv:1511.05644}, 2015.

\bibitem{Mathieu:2016}
M.~F. Mathieu, J.~J. Zhao, J.~Zhao, A.~Ramesh, P.~Sprechmann, and Y.~LeCun,
  ``Disentangling factors of variation in deep representation using adversarial
  training,'' in \emph{Advances in Neural Information Processing Systems},
  2016, pp. 5040--5048.

\bibitem{lample2017fader}
G.~Lample, N.~Zeghidour, N.~Usunier, A.~Bordes, L.~Denoyer \emph{et~al.},
  ``Fader networks: Manipulating images by sliding attributes,'' in
  \emph{Advances in Neural Information Processing Systems}, 2017.

\bibitem{Creswell:2017}
A.~Creswell, A.~A. Bharath, and B.~Sengupta, ``Conditional autoencoders with
  adversarial information factorization,'' \emph{arXiv preprint
  arXiv:1711.05175}, 2017.

\bibitem{Tzeng:2017}
E.~Tzeng, J.~Hoffman, K.~Saenko, and T.~Darrell, ``Adversarial discriminative
  domain adaptation,'' in \emph{Computer Vision and Pattern Recognition},
  vol.~1, no.~2, 2017, p.~4.

\bibitem{Shen:2017}
J.~Shen, Y.~Qu, W.~Zhang, and Y.~Yu, ``Adversarial representation learning for
  domain adaptation,'' \emph{arXiv preprint arXiv:1707.01217}, 2017.

\bibitem{Wang:2018}
Y.~Wang, T.~Koike-Akino, and D.~Erdogmus, ``Invariant representations from
  adversarially censored autoencoders,'' \emph{arXiv preprint
  arXiv:1805.08097}, 2018.

\bibitem{Kingma:2013}
D.~P. Kingma and M.~Welling, ``Auto-encoding variational {Bayes},'' \emph{arXiv
  preprint arXiv:1312.6114}, 2013.

\bibitem{Goodfellow:2014}
I.~Goodfellow, J.~Pouget-Abadie, M.~Mirza, B.~Xu, D.~Warde-Farley, S.~Ozair,
  A.~Courville, and Y.~Bengio, ``Generative adversarial nets,'' in
  \emph{Advances in Neural Information Processing Systems}, 2014.

\bibitem{Lawhern:2016}
V.~J. Lawhern, A.~J. Solon, N.~R. Waytowich, S.~M. Gordon, C.~P. Hung, and
  B.~J. Lance, ``\uppercase{EEGn}et: {A} compact convolutional network for
  \uppercase{EEG}-based brain-computer interfaces,'' \emph{arXiv preprint
  arXiv:1611.08024}, 2016.

\bibitem{Schirrmeister:2017}
R.~T. Schirrmeister, J.~T. Springenberg, L.~D.~J. Fiederer, M.~Glasstetter,
  K.~Eggensperger, M.~Tangermann, F.~Hutter, W.~Burgard, and T.~Ball, ``Deep
  learning with convolutional neural networks for \uppercase{EEG} decoding and
  visualization,'' \emph{Human Brain Mapping}, vol.~38, no.~11, pp. 5391--5420,
  2017.

\bibitem{Bashivan:2015}
P.~Bashivan, I.~Rish, M.~Yeasin, and N.~Codella, ``Learning representations
  from \uppercase{EEG} with deep recurrent-convolutional neural networks,''
  \emph{arXiv preprint arXiv:1511.06448}, 2015.

\bibitem{Luo:2018}
Y.~Luo and B.-L. Lu, ``\uppercase{EEG} data augmentation for emotion
  recognition using a conditional {Wasserstein} \uppercase{GAN},'' in
  \emph{International Conference of the IEEE Engineering in Medicine and
  Biology Society}, 2018.

\bibitem{Hartmann:2018}
K.~G. Hartmann, R.~T. Schirrmeister, and T.~Ball, ``\uppercase{EEG-GAN}:
  Generative adversarial networks for electroencephalograhic (\uppercase{EEG})
  brain signals,'' \emph{arXiv preprint arXiv:1806.01875}, 2018.

\bibitem{Sohn:2015}
K.~Sohn, H.~Lee, and X.~Yan, ``Learning structured output representation using
  deep conditional generative models,'' in \emph{Advances in Neural Information
  Processing Systems}, 2015, pp. 3483--3491.

\bibitem{Goldberger:2000}
A.~L. Goldberger, L.~A. Amaral, L.~Glass, J.~M. Hausdorff, P.~C. Ivanov, R.~G.
  Mark, J.~E. Mietus, G.~B. Moody, C.-K. Peng, and H.~E. Stanley, ``Physiobank,
  physiotoolkit, and physionet: components of a new research resource for
  complex physiologic signals,'' \emph{Circulation}, vol. 101, no.~23, pp.
  e215--e220, 2000.

\bibitem{Schalk:2004}
G.~Schalk, D.~J. McFarland, T.~Hinterberger, N.~Birbaumer, and J.~R. Wolpaw,
  ``\uppercase{BCI}2000: a general-purpose brain-computer interface
  (\uppercase{BCI}) system,'' \emph{IEEE Transactions on Biomedical
  Engineering}, vol.~51, no.~6, pp. 1034--1043, 2004.

\bibitem{Pfurtscheller:2001}
G.~Pfurtscheller and C.~Neuper, ``Motor imagery and direct brain-computer
  communication,'' \emph{Proceedings of the IEEE}, vol.~89, no.~7, pp.
  1123--1134, 2001.

\bibitem{Tokui:2015}
S.~Tokui, K.~Oono, S.~Hido, and J.~Clayton, ``Chainer: a next-generation open
  source framework for deep learning,'' in \emph{Proceedings of Workshop on
  Machine Learning Systems in the 29th Annual Conference on Neural Information
  Processing Systems}, 2015.

\bibitem{Kingma:2015}
D.~P. Kingma and J.~B. Adam, ``A method for stochastic optimization,'' in
  \emph{International Conference on Learning Representations}, vol.~5, 2015.

\end{thebibliography}


\end{document}